\documentclass{bmvc2k}
\usepackage{subfigure}
\usepackage{pifont}
\usepackage{amssymb}
\usepackage{amsmath}
\usepackage{amsfonts}
\usepackage{booktabs}
\usepackage{etoolbox,siunitx}
\newcommand{\cmark}{\ding{51}}%
\newcommand{\xmark}{\ding{53}}%
\renewcommand{\vec}[1]{\mathbf{#1}}
\robustify\bfseries

\title{An Efficient End-to-End Neural Model for Handwritten Text Recognition}

\addauthor{Arindam Chowdhury}{chowdhury.arindam1@tcs.com}{1}
\addauthor{Lovekesh Vig}{lovekesh.vig@tcs.com}{1}

\addinstitution{TCS Innovation Labs\\Delhi}
\graphicspath{{./images/}}
\runninghead{Chowdhury, Vig}{An Efficient End-to-End Neural Model for Handwritten}

\begin{document}

\maketitle

\begin{abstract}
Offline handwritten text recognition from images is an important problem for enterprises attempting to digitize large volumes of handmarked scanned documents/reports. Deep recurrent models such as Multi-dimensional LSTMs \cite{graves2009offline} have been shown to yield superior performance over traditional Hidden Markov Model based approaches that suffer from the Markov assumption and therefore lack the representational power of RNNs. In this paper we introduce a novel approach that combines a deep convolutional network with a recurrent Encoder-Decoder network to map an image to a sequence of characters corresponding to the text present in the image. The entire model is trained end-to-end using Focal Loss \cite{lin2017focal}, an improvement over the standard Cross-Entropy loss that addresses the class imbalance problem, inherent to text recognition. To enhance the decoding capacity of the model, Beam Search algorithm is employed which searches for the best sequence out of a set of hypotheses based on a joint distribution of individual characters. Our model takes as input a downsampled version of the original image thereby making it both computationally and memory efficient. The experimental results were benchmarked against two publicly available datasets, IAM and RIMES. We surpass the state-of-the-art word level accuracy on the evaluation set of both datasets by $3.5\%$ \& $1.1\%$, respectively.  

\end{abstract}

\section{Introduction}

Handwritten Text Recognition (HTR) has been a major research problem for several decades \cite{bunke2003recognition} \cite{vinciarelli2002survey} and has gained recent impetus due to the potential value that can be unlocked from extracting the data stored in handwritten documents and exploiting it via modern AI systems. Traditionally, HTR is divided into two categories: offline and online recognition. In this paper, we consider the offline recognition problem which is considerably more challenging as, unlike the online mode which exploits attributes like stroke information and trajectory in addition to the text image, offline mode has only the image available for feature extraction. 

Historically, HTR has been formulated as a sequence matching problem: a sequence of features extracted from the input data is matched to an output sequence composed of characters from the text, primarily using \textit{Hidden Markov Models} ( HMM ) \cite{el1999hmm}\cite{marti2001using}. However, HMMs fail to make use of the context information in a text sequence, due to the Markovian assumption that each observation depends only on the current state. This limitation was addressed by the use of \textit{Recurrent Neural Networks} ( RNN ) which encode the context information in the hidden states. Nevertheless, the use of RNN was limited to scenarios in which the individual characters in a sequence could be segmented, as the RNN objective functions require a separate training signal at each timestep. Improvements were proposed in form of models that have a hybrid architecture combining HMM with RNN \cite{bourlard2012connectionist} \cite{bengio1999markovian}, but major breakthrough came in \cite{graves2009novel} which proposed the use of \textit{Connectionist Temporal Classification} ( CTC ) \cite{graves2006connectionist} in combination with RNN. CTC allows the network to map the input sequence directly to a sequence of output labels, thereby doing away with the need of segmented input.  
  
The performance of RNN-CTC model was still limited as it used handcrafted features from the image to construct the input sequence to the RNN. \textit{Multi-Dimensional Recurrent Neural Network} (MDRNN) \cite{graves2009offline} was proposed as the first end-to-end model for HTR. It uses a hierarchy of multi-dimensional RNN layers that process the input text image along both axes thereby learning long term dependencies in both directions. The idea is to capture the spatial structure of the characters along the vertical axis while encoding the sequence information along the horizontal axis. Such a formulation is computationally expensive as compared to standard convolution operations which extract the same visual features as shown in \cite{puigcerver2017multidimensional}, which proposed a composite architecture that combines a \textit{Convolutional Neural Network} ( CNN ) with a deep one-dimensional RNN-CTC model and holds the current state-of-the-art performance on standard HTR benchmarks. 

In this paper, we propose an alternative approach which combines a convolutional network as a feature extractor with two recurrent networks on top for sequence matching. We use the RNN based \textit{Encoder-Decoder} network \cite{cho2014learning} \cite{sutskever2014sequence}, that essentially performs the task of generating a target sequence from a source sequence and has been extensively employed for Neural Machine Translation ( NMT ). Our model incorporates a set of improvements in architecture, training and inference process in the form of Batch \& Layer Normalization, Focal Loss and Beam Search to name a few. Random distortions were introduced in the inputs as a regularizing step while training. Particularly, we make the following key contributions:

\begin{figure}[!b]
\centering
\includegraphics[scale=0.319]{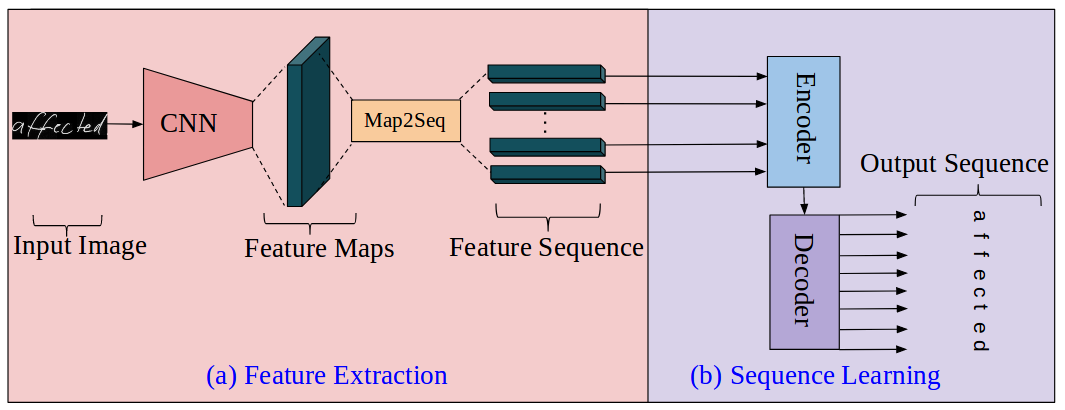}
\caption{Model Overview: (a) represents the generation of feature sequence from convolutional feature maps and (b) shows the mapping of the visual feature sequence to a string of output characters.}
\label{fig:blkdiag}
\end{figure}

\begin{itemize}
\item We present an end-to-end neural network architecture composed of convolutional and recurrent networks to perform efficient offline HTR on images of text lines.

\item We demonstrate that the Encoder-Decoder network with \textit{Attention} provides significant boost in accuracy as compared to the standard RNN-CTC formulation for HTR. 

\item We show that a reduction of $61\%$ in computations and $17\%$ in memory consumption can be achieved by downsampling the input images to almost a sixteenth of the original size, without compromising with the overall accuracy of the model.
\end{itemize}

\section{Proposed Method} \label{sec:promet}

Our model is composed of two connectionist components$:$ $(a)$ \textit{Feature Extraction} module that takes as input an image of a line of text to extract visual features and $(b)$ \textit{Sequence Learning} module that maps the visual features to a sequence of characters. 
A general overview of the model is shown in Figure \ref{fig:blkdiag}. It consists of differentiable neural modules with a seamless interface, allowing fast and efficient end-to-end training.

\subsection{Feature Extraction}

\begin{figure}[ht]
\centering
\includegraphics[scale=0.3]{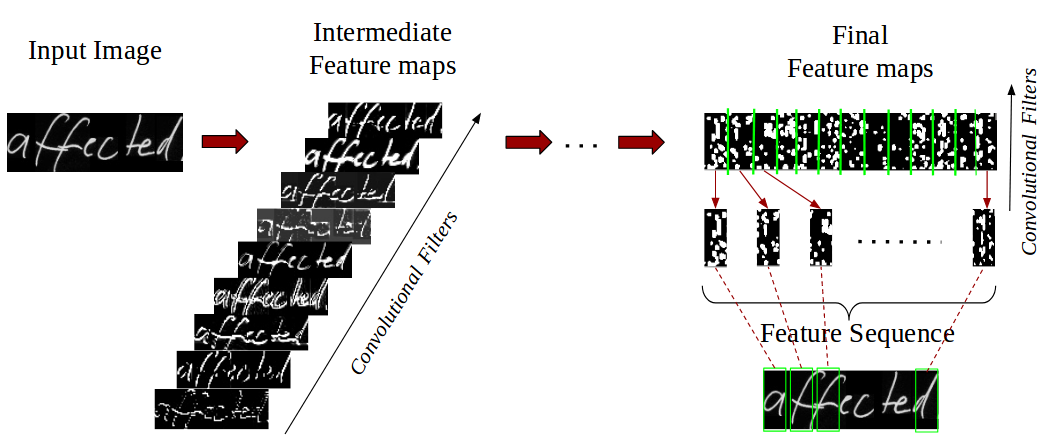}
\caption{Visualization of feature sequence generation process and the possible Receptive fields of the feature vectors. Intermediate feature maps are stacked depth-wise in correspondence with the convolutional filters that generate them while the final feature maps are stacked row-wise.}
\label{fig:map2seq}
\end{figure}

Convolutional Networks have proven to be quite effective in extracting rich visual features from images, by automatically learning a set of non-linear transformations, essential for a given task. Our aim was to generate a sequence of features which would encode local attributes in the image while preserving the spatial organization of the objects in it. Towards this end, we use a standard CNN ( without the fully-connected layers ) to transform the input image into a dense stack of feature maps. A specially designed \textit{Map-to-Sequence} \cite{shi2017end} layer is put on top of the CNN to convert the feature maps into a sequence of feature vectors, by depth-wise detaching columns from it. It means that the $i$-th feature vector is constructed by concatenating the $i$-th columns of all the feature maps. Due to the translational invariance of convolution operations, each column represents a vertical strip in the image ( termed as \textit{Receptive field} ), moving from left to right, as shown in Figure \ref{fig:map2seq}. Before feeding to the network, all the images are scaled to a fixed height while the width is scaled maintaining the aspect ratio of the image. This ensures that all the vectors in the feature sequence conform to the same dimensionality without putting any restriction on the sequence length. 

\subsection{Sequence Learning}

The visual feature sequence extracted by the CNN is used to generate a target sequence composed of character tokens corresponding to the text present in the image. Our main aim, therefore, was to map a variable length input sequence into another variable length output sequence by learning a suitable relationship between them. In the Encoder-Decoder framework, the model consists of two recurrent networks, one of which constructs a compact representation based on its understanding of the input sequence while the other uses the same representation to generate the corresponding output sequence.
 
\begin{figure}[ht]
\centering
\includegraphics[scale=0.35]{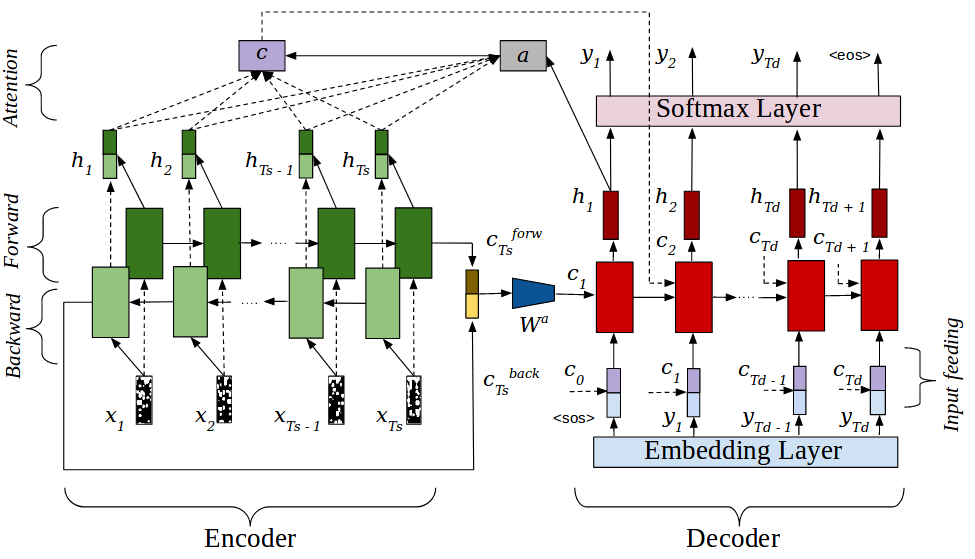}
\caption{RNN \textit{Encoder-Decoder} Network with Attention layer. Encoder is a bidirectional LSTM whose outputs are concatenated at each step while decoder is an unidirectional LSTM with a \textit{Softmax} layer on top. The character inputs are sampled from an embedding layer.}
\label{fig:seq2seq}
\end{figure}

The encoder takes as input, the source sequence $\vec{x} = ( x_1, \dots ,x_{T_s} )$, where $T_s$ is the sequence length, and generates a context vector $c$, representative of the entire sequence. This is achieved by using an RNN such that, at each timestep $t$, the hidden state $h_t = g( x_t, h_{t-1} )$ and finally, $c = s( h_1, \dots, h_{T_s} )$, where $g$ and $s$ are some non-linear functions. Such a formulation using a basic RNN cell is quite simple yet proves to be ineffective while learning even slightly long sequences due to the vanishing gradient effect \cite{hochreiter2001gradient}\cite{bengio1994learning} caused by repeated multiplications of gradients in an unfolded RNN. Instead, we use the \textit{Long Short Term Memory} ( LSTM )\cite{hochreiter1997long} cells, for their ability to better model and learn long-term dependencies due to the presence of a memory cell $c \in \mathbb{R}^n$. 
The final cell state $c_{T_s}$ is used as the context vector of the input sequence. In spite of its enhanced memory capacity, LSTM cells are unidirectional and can only learn past context. To utilize both forward and backward dependencies in the input sequence, we make the encoder \textit{bidirectional} \cite{schuster1997bidirectional} , by combining two LSTM cells, which process the sequence in opposite directions, as shown in Figure \ref{fig:seq2seq}. The output of the two cells, forward $\overrightarrow{h}_t$ and backward $\overleftarrow{h}_t$ are concatenated at each timestep, to generate a single output vector $h_t = [\overrightarrow{h}_t ; \overleftarrow{h}_t]$. Similarly, final cell state is formed by concatenating the final forward and backward states $c_{T_s} = [\overrightarrow{c}_{T_s} ; \overleftarrow{c}_{T_s}]$. 

The context vector $c_{T_s}$ is fed to a second recurrent network, called decoder which is used to generate the target sequence. Following an affine transformation, $c_1 = W^ac_{T_s}$, where $W^a$ is the transformation matrix, $c_1$ is used to initialize the cell state of the decoder. Unlike the encoder, decoder is unidirectional as its purpose is to generate, at each timestep $t$, a token $y_t$ of the target sequence, conditioned on $c_1$ and its own previous predictions $\{y_1, \dots, y_{t-1}\}$. Basically, it learns a conditional probability distribution $p(\vec{y}) = \prod_{t=1}^{T_d} p(y_t | \{y_1, \dots,y_{t-1}\}, c_1 )$ over the target sequence $\vec{y} = \{y_1,\dots,y_{T_d}\}$, where $T_d$ is the sequence length. Using an RNN, each conditional is modeled as $p(y_t | \{y_1, \dots,y_{t-1}\}, c_1 ) = Softmax( g(y_{t-1}, h_{t-1}, c_1) )$ , where $g$ is a non-linear function and $h_{t-1}$ is the RNN hidden state. As in case of the encoder, we employ an LSTM cell to implement $g$.  

The above framework proves to be quite efficient in learning a \textit{sequence-to-sequence} mapping but suffers from a major drawback nonetheless. The context vector that forms a link between the encoder and the decoder often becomes an information bottleneck\cite{cho2014learning}. Especially for long sequences, the context vector tends to forget essential information that it saw in the first few timesteps. \textit{Attention} models are an extension to the standard encoder-decoder framework in which the context vector is modified at each timestep based on the similarity of the previous decoder hidden state $h^{decoder}_{t-1}$ with the sequence of annotations $\{h^{encoder}_{1},\dots,h^{encoder}_{T_s}\}$ generated by the encoder, for a given input sequence. As we use a bidirectional encoder, Bahdanau \cite{bahdanau2014neural} attention mechanism becomes a natural choice for our model. The context vector at the $i$-th decoder timestep is given by, 
\begin{align*}
c_i = \Sigma_{j=1}^{T_s} \alpha_{ij}h_j^{encoder}
\end{align*}
The weight $\alpha_{ij}$ for each $h_j^{encoder}$ is given as,
\begin{align*}
\alpha_{ij} &= \frac{\exp(e_{ij})}{\Sigma_{k=1}^{T_s}\exp(e_{ik})}\\
where, \quad  e_{ij} &= a(h^{decoder}_{i-1},h^{encoder}_{j})
\end{align*}
Here, $a$ is a feedforward network trained along with the other components.

Therefore, the context vector is modified as an weighted sum of the input annotations, where the weights measure how similar the output at position $i$ is with the input around position $j$. Such a formulation helps the decoder to learn local correspondence between the input and output sequences in tandem with a global context, which becomes especially useful in case of longer sequences. Additionally, we incorporate the attention \textit{input feeding} approach used in Luong \cite{luong2015effective} attention mechanism in which the context vector from previous timestep is concatenated with the input of the current timestep. It helps in building a local context, further augmenting the predictive capacity of the network. 

We train the model by minimizing a cumulative \textit{categorical cross-entropy} ( CE ) loss calculated independently for each token in a sequence and then summed up. For a target sequence $\vec{y} = \{ y_1, \dots, y_{T_d} \}$, the loss is defined as $CE(\vec{y}) = -\Sigma_{t=1}^{T_d} \log(p(y_t))$ where $p(y_t)$ is the probability of true class at timestep $t$.  The input to the decoder at each timestep is an embedding vector, from a learnable embedding layer, corresponding to the gold prediction from previous step, until the \textit{end-of-sequence} or $<$eos$>$ token is emitted. At this point, a step of gradient descent is performed across the recurrent network using \textit{Back Propagation Through Time} ( BPTT ) followed by back propagation into the CNN to update the network parameters.

Although CE loss is a powerful measure of network performance in a complex multi-class classification scenario, it often suffers from class imbalance problem. In such a situation, the CE loss is mostly composed of the easily classified examples which dominate the gradient. \textit{Focal Loss} \cite{lin2017focal} addresses this problem by assigning suitable weights to the contribution of each instance in the final loss. It is defined as $FL(p) = -(1-p)^\gamma \log(p)$, where $p$ is the true-class probability and $\gamma$ is a tunable focusing parameter. Such a formulation ensures that the easily classified examples get smaller weights than the hard examples in the final loss, thereby making larger updates for the hard examples. Our primary motivation to use focal loss arises from the fact that, in every language, some characters in the alphabet have higher chances of occurring in regular text than the rest. For example, vowels occur with a higher frequency in English text than a character like \textit{z}. Therefore, to make our model robust to such an inherent imbalance, we formulate our sequence loss as $FL(\vec{y}) = -\Sigma_{t=1}^{T_d} (1-p(y_t))^\gamma \log(p(y_t))$. We found that $\gamma = 2$ worked best for our model.   

To speed up training, we employ mini-batch gradient descent. Here, we optimize a batch loss which is a straightforward extension of the sequence loss, calculated as 
\begin{align*}
L = - \frac{1}{M} \Sigma_{i=1}^M \Sigma_{t=1}^{T_d} (1-p(y_{it}))^\gamma \log(p(y_{it}))
\end{align*}
where $M$ is the batch size and $y_{it}$ represents the $t$-th timestep of the $i$-th instance of the batch.

For any sequence model, the simplest approach for inference is to perform a \textit{Greedy Decoding} ( GD ) which emits, at each timestep, the class with the highest probability from the softmax distribution, as the output at that instance. GD operates with the underlying assumption that the best sequence is composed of the most likely tokens at each timestep, which may not necessarily be true. A more refined decoding algorithm is the \textit{Beam Search} which aims to find the best sequence by maximizing the joint distribution,
\begin{align*}
p(y_1, y_2, \dots, y_{T_d}) = p(y_1) \times p(y_2|y_1) \times p(y_3|\{y_1,y_2\}) \times \dots \times p(y_{T_d} | \{y_1, y_2, \dots, y_{T_d - 1}\}) 
\end{align*} 
over a set of hypotheses, known as the beam. The algorithm selects top-$K$ classes, where $K$ is the beam size, at the first timestep and obtains an output distribution individually for each of them at the next timestep. Out of the $K \times N$ hypotheses, where $N$ is the output vocabulary size, the top-$K$ are chosen based on the product $p(y_1) \times p(y_2|y_1)$. This process is repeated till all the $K$ rays in the beam emit the $<$eos$>$ token. The final output of the decoder is the ray having the highest value of  $p(y_1, y_2, \dots, y_{T_d})$ in the beam.

\section{Implementation Details} \label{sec:impdet}


\subsection{Image Preprocessing}

The input to our system are images that contain a line of handwritten text which may or may not be a complete sentence. The images have a single channel with $256$ intensity levels. We invert the images prior to training so that the foreground is composed of higher intensity on a dark background, making it slightly easier for the CNN activations to learn. We also scale down the input images from an average height of $128$ pixels to $32$ pixels while the width is scaled maintaining the aspect ratio of the original image to reduce computations and memory requirements as shown in Table \ref{tab:table2}. As we employ minibatch training, uniformity in dimensions is maintained in a batch by padding the images with background pixels on both left and right to match the width of the widest image in the batch. In preliminary experiments, our model had shown a tendency to overfit on the training data. To prevent such an outcome, as a further regularization step, we introduced random distortions \cite{puigcerver2017multidimensional} in the training images, so that, ideally, in every iteration the model would process a previously unseen set of inputs. Every training batch is subjected to a set of four operations viz. translation, rotation, shear and scaling. Parameters for all the operations are sampled independently from a Gaussian distribution. The operations and the underlying distribution were chosen by observing a few examples at the beginning of experimentations and were fixed then on.

\begin{table}[!t]
\parbox{.45\linewidth}{
\centering
    \caption{Network configuration}
    \label{tab:table1}
    \begin{tabular}{|l|c|} 
	  \hline      
        & Layers\\
	    Configuration & $ 1 - 2 - 3 - 4 - 5 - 6 - 7 $\\      
      \hline
      Conv. filters & 16 32 64 64 128 128 128 \\ 
      Maxpool( $2$x$2$ ) & \cmark - \cmark - \xmark - \xmark - \xmark - \xmark - \xmark \\
      Maxpool( $2$x$1$ ) & \xmark - \xmark - \xmark - \xmark - \cmark - \cmark - \xmark \\
      \hline
    \end{tabular}
}
\hfill
\parbox{.45\linewidth}{
\centering
    \caption{Effect of Image Downsampling on Model Performance while Training}
    \label{tab:table2}
    \begin{tabular}{|c|c|c|}
	  \hline      
      & Computations  & Memory\\
	  Size & ( Tflops ) & ( GB )\\      
      \hline
      $128 \times W$ & $1.5 \times 10^4$ & $9.5$\\ 
      $32 \times W'$ & $5.9 \times 10^3$ & $7.9$\\
      \hline
    \end{tabular}
}
\end{table}

\subsection{Convolutional Network}

Our model consists of seven convolutional ( conv ) layers stacked serially, with \textit{Leaky ReLU} \cite{maas2013rectifier} activations. The first six layers use a kernel size of $3$x$3$ pixels with $1$ pixel wide input padding while the final layer uses a kernel size of $2$x$2$ pixels without input padding. Kernel strides are of $1$ pixel in both vertical and horizontal directions. Activations of the conv layers are \textit{Batch Normalized} \cite{ioffe2015batch}, to prevent internal covariate shift and thereby speed up training, before propagating to the next layer. Pooling operations are performed on the activations of certain conv layers to reduce the dimensionality of the input. A total of four max-pooling layers are used in our model, two of which have a kernel size of $2$x$1$ to preserve the horizontal spatial distribution of text and the rest use standard $2$x$2$ non-overlapping kernels. Table \ref{tab:table1} shows the network configuration used in each conv layer.   

\subsection{RNN Encoder-Decoder}

Encoder \& decoder use LSTM cells with $256$ hidden units. We allow both networks to extend to a depth of $2$ layers to enhance their learning capacity. \textit{Residual connections} \cite{kim2017residual} are created to facilitate gradient flow across the recurrent units to the layers below. Further, we use \textit{dropout} \cite{pham2014dropout} along depth connections to regularize the network, without modifying the recurrent connections, thereby preserving the network's capacity to capture long-term dependencies. To prevent covariate shift due to minibatch training, the activities of the cell neurons are \textit{Layer Normalized} \cite{ba2016layer}, which proved to be quite effective in stabilizing the hidden state dynamics of the network. For the final prediction we apply a linear transformation $W \in \mathbb{R}^{256xN}$ on the RNN predictions, where $N$ is the output vocabulary size, to generate the logits. \textit{Softmax} operation is performed on the logits to define a probability distribution over the output vocabulary at each timestep. 

\subsection{Training \& Inference}

 In our experiments, while training, the batch size is set to $16$. We use \textit{Adam} \cite{kingma2014adam} algorithm as the optimizer with a learning rate of $0.001$. The model was trained till the best validation accuracy, achieved after $30$ epochs. 
For inference, we use a beam size equal to the number of classes in the output. 

\section{Dataset}\label{sec:data}

We use the following publicly available datasets to evaluate our method.\\
\textbf{IAM Handwriting Database} v$3.0$ ( English ) \cite{marti2002iam} is composed of $1539$ pages of text, written by $657$ different writers and partitioned into writer-independent training, validation and test sets of $6161$, $940$, $1861$ segmented lines, respectively. The line images have an average height of $124$ pixels and average width of $1751$ pixels. There are $79$ different characters in the database, including whitespace.\\
\textbf{RIMES Database} ( French ) \cite{augustin2006rimes} has $12,723$ scanned pages of mails handwritten by $1300$ people. The dataset consists of $11333$ segmented lines for training and $778$ for testing. Original database doesn't provide a separate validation set and therefore we randomly sampled $10$\% of the total training lines for validation. Final partition of the dataset contains $10203$ training lines, $1130$ validation lines and $778$ test lines. Average width of the images is $1658$ pixels and average height is $113$ pixels. There are $99$ different characters in the dataset.   

%
%

\section{Experiments} \label{sec:res}
We evaluate our model on the evaluation partition of both datasets using mean \textit{Character Error Rate} ( CER ) and mean \textit{Word Error Rate} ( WER ) as performance metrics determined as the mean over all text lines. They are defined as,

\begin{align*}
\begin{split}
& CER = \frac{\text{Number of characters in a sentence wrongly classified}}{\text{Total characters in a sentence}}\\\\
& WER = \frac{\text{Number of words in a sentence spelled incorrectly}}{\text{Total words in a sentence}}
\end{split}
\end{align*}

Our experiments were performed using an Nvidia Tesla K40 GPU. The inference time for the model is $1.5$ seconds. 

\subsection{Results}

Table \ref{tab:table3} shows the effect of Layer Normalization ( LN ), Focal Loss and Beam Search on the base model. LN improved the performance of the base model by around $3\%$. The use of Focal Loss also increased the accuracy by $1-3\%$ but major improvement was achieved by replacing greedy decoding with beam search which boosted the model accuracy by $4-5\%$. 

\begin{table}[!b]
  \sisetup{parse-numbers=false}
  \begin{center}
    \caption{Effect of Layer Normalization, Focal Loss \& Beam Search on Model Performance}
    \label{tab:table3}
	\noindent\begin{tabular}{lcccc}
	\hline
	& \multicolumn{2}{c}{IAM} & \multicolumn{2}{c}{RIMES} \\
	\midrule
	System & {CER(\%)} & {WER(\%)}  & {CER(\%)} & {WER(\%)} \\
	\hline
    Baseline & 17.4 & 25.5 & 12.0 & 19.1 \\
    + LN & 13.1 & 22.9 & 9.7 & 15.8 \\
    + LN + Focal Loss & 11.4 & 21.1 & 7.3 & 13.5 \\     
    + LN + Focal Loss + Beam Search & \textbf{8.1} & \textbf{16.7} & \textbf{3.5} & \textbf{9.6} \\
	\hline
	\end{tabular}
  \end{center}
\end{table}

\subsection{Comparison with the state-of-the-art}

We provide a comparison of the accuracy of our method with previously reported algorithms in Table \ref{tab:table4} and a comparison of the efficiency, in terms of maximum GPU memory consumption and number of trainable parameters, with the state-of-the-art in Table \ref{tab:table5}.

\begin{table}[htbp!]
  \sisetup{parse-numbers=false}
  \begin{center}
    \caption{Comparison with previous methods in terms of accuracy}
    \label{tab:table4}
	\noindent\begin{tabular}{lcccc}
	\hline
	& \multicolumn{2}{c}{IAM} & \multicolumn{2}{c}{RIMES} \\
	\midrule
	Methods & {CER(\%)} & {WER(\%)}  & {CER(\%)} & {WER(\%)} \\
	\hline
    2DLSTM \cite{graves2009offline}, reported by \cite{puigcerver2017multidimensional} & 8.3 & 27.5 & 4.0 & 17.7\\
    CNN-1DLSTM-CTC \cite{puigcerver2017multidimensional} & 6.2 & 20.2 & 2.6 & 10.7\\
	Our method & 8.1 & \textbf{16.7} & 3.5 & \textbf{9.6} \\
	\hline
	\end{tabular}
  \end{center}
\end{table}

\begin{table}[!h]
  \begin{center}
    \caption{Comparison with state-of-the-art in terms of efficiency }
    \label{tab:table5}
    \begin{tabular}{lccc} 
	  \hline      
      Methods & Memory ( GB ) & \# of Parameters ( Mi )\\
      \hline
      CNN-1DRNN-CTC \cite{puigcerver2017multidimensional} & 10.5 & 9.3\\ 
      Our method & \textbf{7.9} & \textbf{4.6} \\
      \hline
    \end{tabular}
  \end{center}
\end{table}

Although we beat the state-of-the-art \cite{puigcerver2017multidimensional} at word level accuracy, our  character level accuracy is slightly lower in comparison. It implies that our model is prone to make additional spelling mistakes in words which have already got mislabeled characters in them, but overall makes fewer spelling mistakes at the aggregate word level. This arises out of the inference behavior of our model, which uses the previous predictions to generate the current output and as a result, a prior mistake can trigger a sequence of future errors. But, higher word accuracy proves that most often, our model gets the entire word in a line correct. Essentially, the model is quite accurate at identifying words but when a mistake does occur, the word level prediction is off by a larger number of characters.  

\section{Summary and Extensions} \label{sec:conc}

We propose a novel framework for efficient handwritten text recognition that combines the merits of two extremely powerful deep neural networks. Our model substantially exceeds performance of all the previous methods on a public dataset and beats them by a reasonable margin on another. While the model performs satisfactorily on standard testing data, we intend to carry out further evaluations to ascertain its performance on completely unconstrained settings, with different writing styles and image quality.
 
An extension to the present method would be to develop a training procedure that would optimize a loss dependant on the correctness of a full sequence instead of a cumulative loss of independent characters, resulting in similar behavior of the model at training and inference. Also, a language model can be incorporated in the training scheme to further augment the performance of the model and correct for mistakes, especially for rare sequences or words.

\bibliography{egbib}

\begin{thebibliography}{28}
\providecommand{\natexlab}[1]{#1}
\providecommand{\url}[1]{\texttt{#1}}
\expandafter\ifx\csname urlstyle\endcsname\relax
  \providecommand{\doi}[1]{doi: #1}\else
  \providecommand{\doi}{doi: \begingroup \urlstyle{rm}\Url}\fi

\bibitem[Augustin et~al.(2006)Augustin, Carr{\'e}, Grosicki, Brodin, Geoffrois,
  and Pr{\^e}teux]{augustin2006rimes}
Emmanuel Augustin, Matthieu Carr{\'e}, Emmanu{\`e}le Grosicki, J-M Brodin,
  Edouard Geoffrois, and Fran{\c{c}}oise Pr{\^e}teux.
\newblock Rimes evaluation campaign for handwritten mail processing.
\newblock In \emph{International Workshop on Frontiers in Handwriting
  Recognition (IWFHR'06),}, pages 231--235, 2006.

\bibitem[Ba et~al.(2016)Ba, Kiros, and Hinton]{ba2016layer}
Jimmy~Lei Ba, Jamie~Ryan Kiros, and Geoffrey~E Hinton.
\newblock Layer normalization.
\newblock \emph{arXiv preprint arXiv:1607.06450}, 2016.

\bibitem[Bahdanau et~al.(2014)Bahdanau, Cho, and Bengio]{bahdanau2014neural}
Dzmitry Bahdanau, Kyunghyun Cho, and Yoshua Bengio.
\newblock Neural machine translation by jointly learning to align and
  translate.
\newblock \emph{arXiv preprint arXiv:1409.0473}, 2014.

\bibitem[Bengio(1999)]{bengio1999markovian}
Yoshua Bengio.
\newblock Markovian models for sequential data.
\newblock \emph{Neural computing surveys}, 2\penalty0 (199):\penalty0 129--162,
  1999.

\bibitem[Bengio et~al.(1994)Bengio, Simard, and Frasconi]{bengio1994learning}
Yoshua Bengio, Patrice Simard, and Paolo Frasconi.
\newblock Learning long-term dependencies with gradient descent is difficult.
\newblock \emph{IEEE transactions on neural networks}, 5\penalty0 (2):\penalty0
  157--166, 1994.

\bibitem[Bourlard and Morgan(2012)]{bourlard2012connectionist}
Herve~A Bourlard and Nelson Morgan.
\newblock \emph{Connectionist speech recognition: a hybrid approach}, volume
  247.
\newblock Springer Science \& Business Media, 2012.

\bibitem[Bunke(2003)]{bunke2003recognition}
Horst Bunke.
\newblock Recognition of cursive roman handwriting: past, present and future.
\newblock In \emph{Document Analysis and Recognition, 2003. Proceedings.
  Seventh International Conference on}, pages 448--459. IEEE, 2003.

\bibitem[Cho et~al.(2014)Cho, Van~Merri{\"e}nboer, Gulcehre, Bahdanau,
  Bougares, Schwenk, and Bengio]{cho2014learning}
Kyunghyun Cho, Bart Van~Merri{\"e}nboer, Caglar Gulcehre, Dzmitry Bahdanau,
  Fethi Bougares, Holger Schwenk, and Yoshua Bengio.
\newblock Learning phrase representations using rnn encoder-decoder for
  statistical machine translation.
\newblock \emph{arXiv preprint arXiv:1406.1078}, 2014.

\bibitem[El-Yacoubi et~al.(1999)El-Yacoubi, Gilloux, Sabourin, and
  Suen]{el1999hmm}
A~El-Yacoubi, Michel Gilloux, Robert Sabourin, and Ching~Y. Suen.
\newblock An hmm-based approach for off-line unconstrained handwritten word
  modeling and recognition.
\newblock \emph{IEEE Transactions on Pattern Analysis and Machine
  Intelligence}, 21\penalty0 (8):\penalty0 752--760, 1999.

\bibitem[Graves and Schmidhuber(2009)]{graves2009offline}
Alex Graves and J{\"u}rgen Schmidhuber.
\newblock Offline handwriting recognition with multidimensional recurrent
  neural networks.
\newblock In \emph{Advances in neural information processing systems}, pages
  545--552, 2009.

\bibitem[Graves et~al.(2006)Graves, Fern{\'a}ndez, Gomez, and
  Schmidhuber]{graves2006connectionist}
Alex Graves, Santiago Fern{\'a}ndez, Faustino Gomez, and J{\"u}rgen
  Schmidhuber.
\newblock Connectionist temporal classification: labelling unsegmented sequence
  data with recurrent neural networks.
\newblock In \emph{Proceedings of the 23rd international conference on Machine
  learning}, pages 369--376. ACM, 2006.

\bibitem[Graves et~al.(2009)Graves, Liwicki, Fern{\'a}ndez, Bertolami, Bunke,
  and Schmidhuber]{graves2009novel}
Alex Graves, Marcus Liwicki, Santiago Fern{\'a}ndez, Roman Bertolami, Horst
  Bunke, and J{\"u}rgen Schmidhuber.
\newblock A novel connectionist system for unconstrained handwriting
  recognition.
\newblock \emph{IEEE transactions on pattern analysis and machine
  intelligence}, 31\penalty0 (5):\penalty0 855--868, 2009.

\bibitem[Hochreiter and Schmidhuber(1997)]{hochreiter1997long}
Sepp Hochreiter and J{\"u}rgen Schmidhuber.
\newblock Long short-term memory.
\newblock \emph{Neural computation}, 9\penalty0 (8):\penalty0 1735--1780, 1997.

\bibitem[Hochreiter et~al.(2001)Hochreiter, Bengio, Frasconi, Schmidhuber,
  et~al.]{hochreiter2001gradient}
Sepp Hochreiter, Yoshua Bengio, Paolo Frasconi, J{\"u}rgen Schmidhuber, et~al.
\newblock Gradient flow in recurrent nets: the difficulty of learning long-term
  dependencies, 2001.

\bibitem[Ioffe and Szegedy(2015)]{ioffe2015batch}
Sergey Ioffe and Christian Szegedy.
\newblock Batch normalization: Accelerating deep network training by reducing
  internal covariate shift.
\newblock \emph{arXiv preprint arXiv:1502.03167}, 2015.

\bibitem[Kim et~al.(2017)Kim, El-Khamy, and Lee]{kim2017residual}
Jaeyoung Kim, Mostafa El-Khamy, and Jungwon Lee.
\newblock Residual lstm: Design of a deep recurrent architecture for distant
  speech recognition.
\newblock \emph{arXiv preprint arXiv:1701.03360}, 2017.

\bibitem[Kingma and Ba(2014)]{kingma2014adam}
Diederik~P Kingma and Jimmy Ba.
\newblock Adam: A method for stochastic optimization.
\newblock \emph{arXiv preprint arXiv:1412.6980}, 2014.

\bibitem[Lin et~al.(2017)Lin, Goyal, Girshick, He, and
  Doll{\'a}r]{lin2017focal}
Tsung-Yi Lin, Priya Goyal, Ross Girshick, Kaiming He, and Piotr Doll{\'a}r.
\newblock Focal loss for dense object detection.
\newblock \emph{arXiv preprint arXiv:1708.02002}, 2017.

\bibitem[Luong et~al.(2015)Luong, Pham, and Manning]{luong2015effective}
Minh-Thang Luong, Hieu Pham, and Christopher~D Manning.
\newblock Effective approaches to attention-based neural machine translation.
\newblock \emph{arXiv preprint arXiv:1508.04025}, 2015.

\bibitem[Maas et~al.(2013)Maas, Hannun, and Ng]{maas2013rectifier}
Andrew~L Maas, Awni~Y Hannun, and Andrew~Y Ng.
\newblock Rectifier nonlinearities improve neural network acoustic models.
\newblock In \emph{Proc. icml}, volume~30, page~3, 2013.

\bibitem[Marti and Bunke(2001)]{marti2001using}
U-V Marti and Horst Bunke.
\newblock Using a statistical language model to improve the performance of an
  hmm-based cursive handwriting recognition system.
\newblock In \emph{Hidden Markov models: applications in computer vision},
  pages 65--90. World Scientific, 2001.

\bibitem[Marti and Bunke(2002)]{marti2002iam}
U-V Marti and Horst Bunke.
\newblock The iam-database: an english sentence database for offline
  handwriting recognition.
\newblock \emph{International Journal on Document Analysis and Recognition},
  5\penalty0 (1):\penalty0 39--46, 2002.

\bibitem[Pham et~al.(2014)Pham, Bluche, Kermorvant, and
  Louradour]{pham2014dropout}
Vu~Pham, Th{\'e}odore Bluche, Christopher Kermorvant, and J{\'e}r{\^o}me
  Louradour.
\newblock Dropout improves recurrent neural networks for handwriting
  recognition.
\newblock In \emph{Frontiers in Handwriting Recognition (ICFHR), 2014 14th
  International Conference on}, pages 285--290. IEEE, 2014.

\bibitem[Puigcerver(2017)]{puigcerver2017multidimensional}
Joan Puigcerver.
\newblock Are multidimensional recurrent layers really necessary for
  handwritten text recognition?
\newblock In \emph{Document Analysis and Recognition (ICDAR), 2017 14th IAPR
  International Conference on}, volume~1, pages 67--72. IEEE, 2017.

\bibitem[Schuster and Paliwal(1997)]{schuster1997bidirectional}
Mike Schuster and Kuldip~K Paliwal.
\newblock Bidirectional recurrent neural networks.
\newblock \emph{IEEE Transactions on Signal Processing}, 45\penalty0
  (11):\penalty0 2673--2681, 1997.

\bibitem[Shi et~al.(2017)Shi, Bai, and Yao]{shi2017end}
Baoguang Shi, Xiang Bai, and Cong Yao.
\newblock An end-to-end trainable neural network for image-based sequence
  recognition and its application to scene text recognition.
\newblock \emph{IEEE transactions on pattern analysis and machine
  intelligence}, 39\penalty0 (11):\penalty0 2298--2304, 2017.

\bibitem[Sutskever et~al.(2014)Sutskever, Vinyals, and
  Le]{sutskever2014sequence}
Ilya Sutskever, Oriol Vinyals, and Quoc~V Le.
\newblock Sequence to sequence learning with neural networks.
\newblock In \emph{Advances in neural information processing systems}, pages
  3104--3112, 2014.

\bibitem[Vinciarelli(2002)]{vinciarelli2002survey}
Alessandro Vinciarelli.
\newblock A survey on off-line cursive word recognition.
\newblock \emph{Pattern recognition}, 35\penalty0 (7):\penalty0 1433--1446,
  2002.

\end{thebibliography}
\end{document}